%% file: main.tex
\documentclass[sigconf]{acmart}

\AtBeginDocument{%
  \providecommand\BibTeX{{%
    \normalfont B\kern-0.5em{\scshape i\kern-0.25em b}\kern-0.8em\TeX}}}

\usepackage{adjustbox}
\usepackage{diagbox}
\usepackage{array}

\makeatletter                   
\def\mdseries@tt{m}             
\makeatother                    
\usepackage[plain]{fancyref}
\usepackage[draft=true]{minted} 
\usepackage{color}
\usepackage{hyperref}           
\hypersetup{
    colorlinks=true,
    linkcolor=blue,
    filecolor=red,      
    urlcolor=magenta,
    breaklinks=true,            
}
\usepackage{breakurl}           

\usepackage{colortbl}

\usepackage{makecell}

\newcolumntype{R}[2]{%
    >{\adjustbox{angle=#1,lap=\width-(#2)}\bgroup}%
    r%
    <{\egroup}%
}
\newcommand*\rot{\multicolumn{1}{|R{90}{2em}|}}
\usepackage{color, colortbl}
\usepackage{multirow}
\usepackage{hhline}

\copyrightyear{2019}
\acmYear{2019}
\setcopyright{rightsretained}
\makeatletter
\renewcommand\@copyrightpermission[1]{In: Proceedings of the Third Workshop on Automated Semantic Analysis of Information in Legal Text (ASAIL 2019), June 21, 2019, Montreal, QC, Canada.}
\makeatother
\acmConference[ASAIL 2019]{Third Workshop on Automated Semantic Analysis of Information in Legal Text}{June 21, 2019}{Montreal, QC, Canada}

\input{SentenceExampleEnv.tex}

\begin{document}

\title{Shift-of-Perspective Identification within Legal Cases}


\input{authors}
\renewcommand{\shortauthors}{Ratnayaka and Rupasinghe, et al.}

\begin{abstract}
\input{abstract.tex}

\end{abstract}

\keywords{semantic analysis, sentiment analysis, natural language processing, information extraction, law}

\maketitle

\input{body.tex}

\bibliographystyle{ACM-Reference-Format}
\bibliography{references}


\end{document}

%% file: SentenceExampleEnv.tex
\usepackage{enumitem}

\newcounter{SentenceCounter}
\newcounter{ExampleCounter}

\newenvironment{Sentence}[1][]{
\refstepcounter{SentenceCounter}
\vspace{0.5mm}
\def\temp{#1}\ifx\temp\empty
    \noindent Sentence~\theExampleCounter.\theSentenceCounter:
\else
    \noindent Sentence~\theExampleCounter.\theSentenceCounter\hspace{1mm}(#1):
\fi
\itshape
}
{
\label{sent\theExampleCounter.\theSentenceCounter}
}

\newenvironment{SentenceExample}[1][]{
\setcounter{SentenceCounter}{0}
\vspace{0.5mm}
\begin{table}[htb]
\centering
\bgroup

\begin{tabular}{|p{0.47\textwidth}|}
\hline
\scriptsize
\vspace{1mm}
\def\temp{#1}
\ifx\temp\empty
    \refstepcounter{ExampleCounter}
    \noindent\textbf{Example \theExampleCounter}
\else
    \noindent\textbf{Example \theExampleCounter\hspace{1mm}(#1)}
\fi
\footnotesize
\begin{itemize}[leftmargin=*]

}{
\end{itemize}
\\\hline
\end{tabular}
\egroup
\label{eg\theExampleCounter}
\end{table}
\vspace{0.5mm}
}

%% file: authors.tex
\author{Gathika Ratnayaka}
\email{gathika.14@cse.mrt.ac.lk}
\affiliation{%
  \department{Department of Computer Science \& Engineering}
  \institution{University of Moratuwa}
  \streetaddress{Bandaranayake Mawatha}
  \city{Moratuwa}
  \state{Sri Lanka}
  \postcode{10400}
}

\author{Thejan Rupasinghe}
\email{thejanrupasinghe.14@cse.mrt.ac.lk}
\affiliation{%
  \department{Department of Computer Science \& Engineering}
  \institution{University of Moratuwa}
  \streetaddress{Bandaranayake Mawatha}
  \city{Moratuwa}
  \state{Sri lanka}
  \postcode{10400}
}

\author{Nisansa de Silva}
\email{nisansaDdS@cse.mrt.ac.lk}
\affiliation{%
  \department{Department of Computer Science \& Engineering}
  \institution{University of Moratuwa}
  \streetaddress{Bandaranayake Mawatha}
  \city{Moratuwa}
  \state{Sri Lanka}
  \postcode{10400}
}

\author{Viraj Salaka Gamage}
\email{viraj.14@cse.mrt.ac.lk}
\affiliation{%
  \department{Department of Computer Science \& Engineering}
  \institution{University of Moratuwa}
  \streetaddress{Bandaranayake Mawatha}
  \city{Moratuwa}
  \state{Sri Lanka}
  \postcode{10400}
}

\author{Menuka Warushavithana}
\email{menuka.14@cse.mrt.ac.lk}
\affiliation{%
  \department{Department of Computer Science \& Engineering}
  \institution{University of Moratuwa}
  \streetaddress{Bandaranayake Mawatha}
  \city{Moratuwa}
  \state{Sri Lanka}
  \postcode{10400}
}

\author{Amal Shehan Perera}
\email{shehan@cse.mrt.ac.lk}
\affiliation{%
  \department{Department of Computer Science \& Engineering}
  \institution{University of Moratuwa}
  \streetaddress{Bandaranayake Mawatha}
  \city{Moratuwa}
  \state{Sri Lanka}
  \postcode{10400}
}

\renewcommand{\shortauthors}{Ratnayaka and Rupasinghe, et al.}

%% file: abstract.tex
Arguments, counter-arguments, facts, and evidence obtained via documents related to previous court cases are of essential need for legal professionals. Therefore, the process of automatic information extraction from documents containing legal opinions related to court cases can be considered to be of significant importance. This study is focused on the identification of sentences in legal opinion texts which convey different perspectives on a certain topic or entity.  
We combined several approaches based on semantic analysis, open information extraction, and sentiment analysis to achieve our objective.
Then, our methodology was evaluated with the help of human judges. The outcomes of the evaluation demonstrate that our system is successful in detecting situations where two sentences deliver different opinions on the same topic or entity. The proposed methodology can be used to facilitate other information extraction tasks related to the legal domain. One such task is the automated detection of counter arguments for a given argument. Another is the identification of opponent parties in a court case.


%% file: body.tex
\input{body-sections/introduction}

\input{body-sections/background}

\input{body-sections/methodology}

\input{body-sections/results.tex}

\input{body-sections/conclusions.tex}


%% file: body-sections/introduction.tex
\section{Introduction}

Documents describing legal opinions related to previous court cases carry a significant importance when it comes to the legal literature. The information presented in these legal opinion texts are used in different capacities such as evidence, arguments, and facts by legal officials in the process of constructing new legal cases~\cite{sugathadasa2018legal}. Therefore, information extraction from legal opinion texts can be considered as an area of significant importance, within the topic of \textbf{automatic information extraction} in the legal domain. In order to perform systematic information extraction from a legal opinion text, a system should be able to interpret the meaning of a given text. In the process of interpreting the meaning of a text, understanding the context can be considered as a major requirement, especially in the legal literature.

Identifying how textual units are related to each other within a machine-readable text is an important task when it comes to interpreting the context. Humans are good at comparing two textual units to determine the way in which those two units are connected. Granting this ability to computers is a major discussion topic in the research related to areas of Natural Language Processing and Artificial Intelligence. A sentence can be considered as a textual unit with significant importance in a text. Therefore, analysis of relationships between sentences can be useful to get a clear picture on the information flow within a text which is made up of a considerable number of sentences.  

Similarly, identifying the types of relationships existing between sentences in legal opinion texts can be used to identify the information flow within a legal case. Within a document describing legal opinions related to a court case, different types of relationships between sentences can be observed such as \textit{elaboration} and \textit{contradiction}. Pairs of sentences can be classified into two major groups based on whether the topics which are being discussed by the two sentences in the sentence pair is the same or not. In other words, the two sentences in a sentence pair may discuss the same topic or they may discuss completely different topics. Even if the two sentences are discussing the same topic, the opinions or views presented in the two sentences on the topic may be different. Consider the following sentence pair taken from \textit{Lee v. United States} \cite{1977lee}. 
  
\begin{SentenceExample}
\item \begin{Sentence}Applying the two-part test for ineffective assistance claims from Strickland v. Washington, 466 U. S. 668, the Sixth Circuit concluded that, while the Government conceded that Lee's counsel had performed deficiently, Lee could not show that he was prejudiced by his attorney's erroneous advice.\end{Sentence}

\item \begin{Sentence}Lee has demonstrated that he was prejudiced by his counsel's erroneous advice.\end{Sentence}
\end{SentenceExample}
 
The above two sentences discuss whether a person named Lee was able to convince that he was prejudiced by his attorney's advice or not. While the first sentence says that \textit{Lee could not show that he was prejudiced by his attorney's advice}, the second sentence contradicts the first sentence by saying that \textit{Lee has demonstrated that he was prejudiced by his counsel's erroneous advice}. Thus, the two sentences provide different opinions on the same topic.
 Contradiction is not a necessary condition in order to classify a pair of sentences as providing different opinions on the same topic. For example, consider Example 2 which consists of two adjacent sentences which are also taken from \textit{Lee v. United States} \cite{1977lee}.

\begin{SentenceExample}
\item \begin{Sentence}Although he has lived in this country for most of his life, Lee is not a United States citizen, and he feared that a criminal conviction might affect his status as a lawful permanent resident.\end{Sentence}

\item \begin{Sentence}His attorney assured him there was nothing to worry about--the Government would not deport him if he pleaded guilty.\end{Sentence}
\end{SentenceExample}
 
It can be seen that both sentences in this example discuss the topic -- the deportation of a person named Lee. Though the two sentences here do not provide contradictory information, they provide two different viewpoints regarding the same topic. It can be seen that the opinions of Lee and his attorney on the possibility of Lee being deported is different. Therefore, when discussing sentences with different opinions on the same topic, not only the sentences providing contradictory information but also the sentences providing multiple viewpoints on the same discussion topic should also be considered. In each of the above two examples, Sentence 1 comes before Sentence 2. From this point onward, the first sentence in a sentence pair will be referred to as the \textit{Target Sentence} and the second sentence as the \textit{Source Sentence}.

An important observation which can be made by considering Example 2 is that the identification of the shift in the viewpoint in that particular occasion is not straightforward. This implicit nature makes the task of identifying sentences which provides different opinions on the same discussion topic even more challenging. At the same time, it can be considered a vital task due to its potential to enhance the capabilities of Information Extraction from Legal Text by facilitating automatic detection of counter-arguments, identification of the stance of a particular party in a court case and to discover multiple viewpoints to analyze or evaluate a particular legal situation.

Hence, the objective of this study is to identify sentences which have different perspectives on the same discussion topic in a given court case. For this study, legal opinion texts related to United States court cases were used.The next section provides details on the previous work which are related to our study. Section 3 describes the methodology followed in this study while the outcomes of the study are discussed in Section 4. Finally we conclude our discussion in Section 5.

%% file: body-sections/background.tex
\section{Related Work}

Computing applications which can be considered to be both efficient and effective are scarce due to the challenges in handling legal jargon\cite{jayawardana2017word,schweighofer1993legal,jayawardana2017semi}. The nature of legal documents employing a vocabulary of mixed origin ranging from Latin to English has been put forward as a reasoning for difficulties of building computing applications for the legal domain~\cite{sugathadasa2018legal}.          

Regardless, there have been some recent attempts to circumvent these problems in the legal domain including  information organization  \cite{jayawardana2017deriving,jayawardana2017semi,jayawardana2017word}, information extraction~\cite{sugathadasa2017synergistic} and information retrieval ~\cite{sugathadasa2018legal}. Going forward, owing to the popularity of knowledge embedding in the literature, several studies have taken up the task of embedding legal jargon in vector spaces~\cite{sugathadasa2017synergistic,nay2016gov2vec}. Further, in the information extraction domain, the study by Gamage et al~\cite{gamage2018Sentiment} attempted to build a sentiment annotator for the legal domain and the study by Ratnayaka et al~\cite{oblie2018identifying} attempted to identify relationships among sentences in legal opinion texts.

Discovering situations where two sentences are providing different opinions on the same topic or entity is an important part when it comes to identifying relationships among sentences \cite{oblie2018identifying}. Contradiction is a sufficient but not a necessary condition in this regard. The study \cite{marneffe2008finding} is focused on finding contradictions in text related to the real world context. In an attempt to define contradiction, the same study\cite{marneffe2008finding} claims that ``contradiction occurs
when two sentences are extremely unlikely to be true simultaneously'' and the study \cite{paul2010summarizing} also agrees on that definition. However, the study \cite{marneffe2008finding} also demonstrates that two sentences can be contradictory while being true simultaneously. These characteristics of contradiction make the process of detecting contradiction relationships more complex. 

In order to become contradictory, two textual units can elaborate not only on the same event but also on the same entity. For example, if one sentence in a sentence pair is saying that a person is a United States citizen while the other sentence is saying that very same person is \textbf{not} a United States citizen, it is obvious that the two sentences are providing contradictory information. Here, the contradictory information is upon a person which can be considered as an entity. Therefore, it is more reasonable to consider that in order to be contradictory, texts must elaborate on the same topic.

In order to detect contradiction, different features based on the properties of text have been considered in the previous studies \cite{marneffe2008finding,paul2010summarizing}. Polarity features and Numeric Mismatches are such commonly used features. The study \cite{marneffe2008finding} empirically claims that the precision of detecting contradiction falls when numeric mismatches are considered.

The structures of the texts also play a vital role when it comes to contradiction detection \cite{marneffe2008finding,paul2010summarizing}. Analysis of text structure is helpful in identifying the common entity or event on which the contradiction is occurring. When the structure of a given sentence is considered, the subject-object relationship plays an important role \cite{ollie-emnlp12,angeli2015leveraging}. Analysis of Typed Dependency Graphs \cite{chen2014fast} is another useful approach to understand the structure of a particular text and to obtain necessary information.

Polarities of the sentences in relation to the sentiments can also play a vital role when it comes to identification of sentence pairs which provide different opinions on the same topic. It can be observed  the seminal RNTN  (Recursive  Neural  Tensor  Network) model \cite{socher2013recursive} which is trained on movie reviews is used in many recent studies \cite{manning2014stanford,socher2013reasoning} which perform sentiment analysis. The trained RNTN model \cite{socher2013recursive} has a bias towards the movie review text\cite{gamage2018Sentiment}. In order to overcome this problem, the study by Gamage et al \cite{gamage2018Sentiment} has proposed a methodology to develop a sentiment annotator for the legal domain using transfer learning and has obtained 6\% increase in accuracy over the original model \cite{socher2013recursive} within the legal domain. 

The study by de Silva et al \cite{pubmed} introduces a new algorithm to calculate the oppositeness of triples that can be extracted from  microRNA research paper abstracts using open information extraction.
As the study proposes a mechanism to detect inconsistencies within paragraphs, we see it as one potential methodology which can be adapted to detect \textit{Shift-in-View} relationship between sentences. However, as the above-mentioned study \cite{pubmed} specifically focuses on discovering inconsistencies in the medical domain, it is needed to adopt the proposed methodology to the legal domain in order to detect shift-in-perspectives in legal opinion texts. From this point onward in this paper, we will refer to the study \cite{pubmed} as the \textbf{PubMed Study}.

In the study \cite{zahri2012exploiting}, discourse relations between sentences have been used to generate clusters of similar sentences within texts. A Support Vector Machine model is used in this study\cite{zahri2012exploiting} to determine the relationships existing between sentences. In the process of Multi-Class classification performed using the SVM Model, the study  \cite{zahri2012exploiting} has defined a class named \textit{Change of Topic} which combines the \textit{Contradiction} and \textit{Change of Perspective} relations as defined in Cross Document Structure Theory (CST) \cite{radev2000common}. The study\cite{zahri2012exploiting} has obtained lower results for \textit{Change of Topic} than other relationship types and it claims that average results are due to lack of significant features which could properly detect \textit{Contradiction} and \textit{Change of Perspective}. CST relations and data from CST bank have also been used to train an SVM model in another study \cite{oblie2018identifying} in order to predict relationships between sentences in the legal domain. Though the study has done improvements to the features in \cite{zahri2012exploiting} and introduced new features which suit the legal domain, the results obtained in relation to the \textit{Contradiction} and \textit{Change of Perspective} relationships as defined in CST \cite{radev2000common} is very low. One possible reason is that the CST Bank\cite{Radev&al.03} data set is made up of sentences from newspaper articles, where the structural and linguistic features may differ from that in the court case transcripts, especially when it comes to relationships such as \textit{Contradiction} and \textit{Change of Perspective}.


In identifying whether two sentences are providing different perspectives or opinions regarding the same topic, it is important to identify whether the two sentences are discussing the same topic. The study \cite{oblie2018identifying} has proposed a successful methodology to identify whether a given two sentences are discussing the same topic or not. In the same study \cite{oblie2018identifying}, five relationships that can be observed between two sentences are defined as shown below. From this point onward we refer to the system proposed in the study \cite{oblie2018identifying} as \textbf{Sentence Relationship Identifier (SRI)}.

 \begin{itemize}

\item \textbf{Elaboration} - One sentence adds more details to the information provided in the preceding sentence or one sentence develops further on the topic discussed in the previous sentence.  
\item \textbf{Redundancy} - Two sentences provide the same information without any difference or additional information.
\item \textbf{Citation} - A sentence provides references relevant to the details provided in the previous sentence.
\item \textbf{Shift-in-View} - Two sentences are providing conflicting information or different opinions on the same topic or entity.
\item \textbf{No Relation} - No relationship can be observed between the two sentences. One sentence discusses a topic which is different from the topic discussed in another sentence.
\end{itemize}

It can be seen that the relationship type \textit{Shift-in-View} defined in SRI study \cite{oblie2018identifying} aligns with the relationship type that is being discussed in this study. It can also be further confirmed by looking at how CST\cite{radev2000common} relationships are adopted in the study\cite{oblie2018identifying}.

\begin{table}[h]
\caption{Adopting CST Relationships \cite{oblie2018identifying}}
\label{table_adopting_cst_relationships}
\begin{center}
\begin{tabular}{|l|l|}
\hline
\thead{\bfseries Definition} & \thead{\bfseries CST Relationships}\\
\hline
Elaboration & \begin{tabular}[c]{@{}l@{}}Paraphrase, Modality, Subsumption, Elaboration,
\\Indirect Speech,
Follow-up, Overlap,
\\Fulfillment, Description, Historical Background,
\\Reader Profile, Attribution
\end{tabular}\\
\hline
Redundancy & Identity\\
\hline
Citation & Citation\\
\hline
Shift-in-View & \begin{tabular}[c]{@{}l@{}}Change of Perspective,Contradiction \\ 
\end{tabular}\\
\hline
No Relation & -\\
\hline

\end{tabular}
\end{center}
\end{table}

As shown in Table \ref{table_adopting_cst_relationships}, the \textit{Shift-in-View} relationship includes both \textit{Contradiction} and \textit{Change of Perspective} relationships as defined in CST \cite{radev2000common}. \textit{Elaboration}, \textit{Redundancy}, \textit{Shift-in-View} or \textit{Citation} relationships defined in the study\cite{oblie2018identifying} suggest that a sentence pair is discussing the same topic while \textit{No Relation} suggests that the two sentences are discussing completely different topics. It has been stated that SRI is able to detect situations where the discussion topic is changed with a considerable accuracy \cite{oblie2018identifying}. However, it is also stated that the proposed methodology is not able to detect situations where two sentences provide different opinions on the same topic.The results obtained in this study \cite{oblie2018identifying} are shown in Table \ref{table_confusion_matrix}.

\newcommand{\shadedCell}[1]{
#1\% 
\cellcolor{gray!#1!white}
}

\begin{table}[h]
\caption{Confusion Matrix from the Sentence Relationship Identifier study \cite{oblie2018identifying}}
\label{table_confusion_matrix}
\begin{center}
\begin{tabular}{|l|c|c|c|c|c|}\hline
 \diagbox[width=8em]{Actual}{Predicted}& \rot{Elaboration} & \rot{No Relation} & \rot{Citation} & \rot{Shift-in-View} & $\Sigma$ \\
 \hline
 Elaboration& \shadedCell{93.9} & \shadedCell{6.1} & \shadedCell{0.0} & \shadedCell{0.0} & 99\\
 \hline
 No Relation& \shadedCell{11.9} & \shadedCell{88.1} & \shadedCell{0.0} & \shadedCell{0.0} & 42\\
 \hline
 Citation & \shadedCell{0.0} & \shadedCell{4.8} & \shadedCell{95.2} & \shadedCell{0.0} & 21\\
 \hline
 Shift-in-View& \shadedCell{100.0} & \shadedCell{0.0} & \shadedCell{0.0} & \shadedCell{0.0} & 3\\
 \hline
$\Sigma$ & 101 & 44 & 20 & 0 & 165\\
 \hline
\end{tabular}
\end{center}
\end{table}

\begin{table}[h]
\caption{Results from Sentence Relationship Identifier study considering Sentence Pairs where Both Judges Agree \cite{oblie2018identifying}}
\label{table_results_comparison_both_agree_original} 
\begin{center}
\begin{tabular}{|c|c|c|c|}
\hline
\thead{\bfseries Discourse Class} & \thead{\bfseries Precision} & \thead{\bfseries Recall} & \thead{\bfseries F-Measure} \\
\hline
Elaboration & 0.921 & 0.939 & 0.930\\
\hline
No Relation & 0.841 & 0.881 & 0.861\\
\hline
Citation & 1.000 & 0.952 & 0.975\\
\hline
Shift-in-View & - & 0 & -\\
\hline

\end{tabular}
\end{center}
\end{table}

It is clear that the machine learning model inside the SRI is not able to detect \textit{Shift-in-View} relationship. However, Table \ref{table_confusion_matrix} shows that the sentences pairs having \textit{Shift-in-View} relationships are detected as \textit{Elaboration}. It can be considered as a positive aspect, as \textit{Elaboration} suggest that both sentences are elaborating on the same topic, which is a necessary condition when detecting sentences providing different perspectives on the same topic or entity as described in other studies \cite{marneffe2008finding,paul2010summarizing} too.


%% file: body-sections/methodology.tex
\section{Methodology}

\subsection{Identifying Sentence Pairs where Both Sentences Discuss the Same Topic}

It is needed to identify whether the two sentences are discussing the same topic in detecting sentence pairs which provide different opinions on the same topic. Therefore, as the first step, we implemented the Sentence Relationship Identifier(SRI) as it is successful in identifying whether two sentences are discussing on the same topic or not \cite{oblie2018identifying}. 

According to the study \cite{oblie2018identifying}, \textit{Elaboration}, \textit{Redundancy},\textit{Citation} and \textit{Shift-in-View} relationships occur when both sentences discuss the same topic. \textit{Shift-in-View} occurs over \textit{Elaboration} when the two sentences provide different opinions on the same topic. 

 We only consider sentence pairs which are detected as having \textit{Elaboration} relationship type in order to identify whether \textit{Shift-in-View} relationship is present. Though \textit{Redundancy} and \textit{Citation} relationship types also suggest that two sentences are discussing the same topic, the sentence pairs detected with those relationship types are not considered. As the \textit{Redundancy} relationship suggests that two sentences provide similar information, there is no possibility of having different perspectives. In \textit{Citation} relationship, one sentence provides evidence or references to confirm the details presented in the other sentence. Thus, it is not probable to have a situation where two sentences provide different perspectives on the same topic.
 
 However, if the machine learning model described in the study \cite{oblie2018identifying} detect a pair of sentences as having \textit{Shift-in-View} relationship, such a pair will be detected as a sentence pair which provides different opinions on the same topic. Confirming the observations of the study  \cite{oblie2018identifying}, SRI did not identify any pair of sentences as having \textit{Shift-in-View} relationship.
 
 \subsection{Filtering Sentences using Transition Words and Phrases}
 
There are \textit{Transition Words} or \textit{Transition Phrases} which suggest that the \textit{Source Sentence} of a sentence pair is elaborating or building up on the \textit{Target Sentence}. In the \textit{Source Sentence} of  Example 3 (which was taken from  \textit{Lee v. United States} \cite{1977lee}), the transition word \textit{"Accordingly"} implies that the \textit{Source Sentence} is being developed while agreeing with the \textit{Target Sentence}.

\begin{SentenceExample}
\item \begin{Sentence} Lee's claim that he would not have accepted a plea had he known it would lead to deportation is backed by substantial and uncontroverted evidence.\end{Sentence}

\item \begin{Sentence}Accordingly we conclude Lee has demonstrated a ``reasonable probability that, but for [his] counsel's errors, he would not have pleaded guilty and would have insisted on going to trial''
\end{Sentence}
\end{SentenceExample}

Therefore, when such a \textit{Transition Word} or \textit{Transition Phrase} is present in the \textit{Source Sentence}, such a sentence pair will be considered as having the \textit{Elaboration} relationship. As a result, such sentence pairs are not processed further for detecting the \textit{Shift-in-View} relationship type. We have implemented this mechanism as a way to increase the precision of the \textit{Shift-in-View} detection approaches. Given below are some Transition Words and Transition Phrases we used.

Transition Words: \textit{thus}, \textit{accordingly}, \textit{therefore}

Transition Phrases: \textit{as a result}, \textit{in such cases}, \textit{because of that}, \textit{in conclusion}, \textit{according to that}

\subsection{Use of Coreferencing}

Prior to checking for linguistic features which imply that the sentence pair is showing \textit{Shift-in-View} relationship, co-referencing is performed on the sentence pair. For coreferencing, Stanford CoreNLP CorefAnnotator (``coref") \cite{clark2015entity} was used. The co-referencing provides a better picture when the same entities are being mentioned in the two sentences using different names\cite{oblie2018identifying}.

\subsection{Analyzing Relationships between Verbs}

The first linguistic approach to detect deviations in opinions expressed in sentences regarding a particular topic is based on verb comparison. Under this approach, verbs are compared using the negation relationship and using adverbial modifiers. 


In this approach, subject-object pairs in the \textit{Target Sentence} is compared with that of the \textit{Source Sentence}. If the subject or object in one sentence is present in the other sentence, the verbs in the sentences are considered. Here, we do not consider verbs which are lemmatized into \textit{"be", "do"} in order to focus only on effective verbs. The Stanford CoreNLP POS Tagger (``pos") \cite{toutanova2003feature} was used in identifying verbs in sentences. After extracting the verbs in two sentences, each verb in \textit{Target sentence} is compared with each verb in \textit{Source sentence} to detect verb pairs with similar meaning.     



\newcounter{para}
\newcommand\feature{\par\refstepcounter{para}\noindent\thepara.\space}

\subsubsection{Determining Verbs which Convey Similar Meanings}

In order to convey a similar meaning, it is not necessary that both verbs are the same. Also, when semantic similarity measures between two verbs are considered, it can be observed that there are verb pairs which have very similar meanings but different semantic similarity scores. For example, if the lemmatized forms of verbs in Example 1 are considered, it can be observed that the verb \textit{demonstrate} in the \textit{Target sentence} and verb \textit{show} in the \textit{source sentence} have similar meanings. Confirming that observation further, a Wu-Palmer similarity score of 1.0 can be obtained for that verb pair. When the lemmatized forms of verbs in two sentences in Example 2 are considered, it can be observed that the word "fear" in the \textit{Target sentence} and "worry" in \textit{Source sentence} are two verbs with similar meanings. However, the Wu-Palmer semantic similarity score between verbs \textit{fear} and \textit{worry} is 0.889. Therefore, it is needed to determine an acceptable threshold based on semantic similarity scores in order to identify verbs with similar meanings.

In order to determine this threshold, we first took 1000 verb pairs from legal opinion texts, whose Wu-Palmer similarity scores are greater than 0.75. As our objective is to identify pairs of verbs with similar meanings, it could be observed that a Wu-Palmer score of 0.75 was a reasonable lower bound as per the precision values. We annotated those 1000 pairs of verbs based on whether a given verb pair actually has two verbs with similar meanings or not. Then we gradually incremented the threshold by 0.1 from 0.75 to 0.95 and observed the precision and recall values as shown in Table \ref{table_different_scores}.

\begin{table*}[h]
\caption{Results Comparison for Different Wu-Palmer, Jiang-Conrath, and Lin Score Thresholds}
\label{table_different_scores} 
\begin{center}
\begin{tabular}{|c|c|c|c|c|c|c|c|c|c|}
\hline
\multirow{2}{*}{\bfseries Score} & \multicolumn{3}{c|}{\bfseries Wu-Palmer} & \multicolumn{3}{c|}{\bfseries Jiang-Conrath} & \multicolumn{3}{c|}{\bfseries Lin} \\
\hhline{~---------}
& \thead{\bfseries Precision} & \thead{\bfseries Recall} & \thead{\bfseries F-Measure} & \thead{\bfseries Precision} & \thead{\bfseries Recall} & \thead{\bfseries F-Measure} & \thead{\bfseries Precision} & \thead{\bfseries Recall} & \thead{\bfseries F-Measure} \\
\hline
0.75 & 45.65\% & 100.00\% & 62.68\% & 70.78\% & 51.54\% & 59.64\% & 57.29\% & 72.37\% & \textbf{63.95\%} \\
\hline
0.80 & 51.39\% & 77.19\% & 61.70\% & 70.31\% & 49.34\% & 57.99\% & 60.39\% & 67.54\% & \textbf{63.77\%} \\
\hline
0.85 & 54.59\% & 69.08\%  & 60.99\% & 71.02\% & 48.90\% & 57.92\% & 64.76\% & 62.06\% & \textbf{63.38\%} \\
\hline
0.86 & 59.34\% & 59.21\% & 59.28\% & 71.02\% & 48.90\% & 57.92\% & 67.15\% & 60.96\% & \textbf{63.91\%} \\

\hline
0.90 & 64.49\% & 49.78\% & 56.19\% & 71.25\% & 48.90\% & 58.00\% & 70.40\% & 53.73\% & \textbf{60.95\%} \\
\hline
0.95 & 72.69\% & 41.45\% & 52.79\% & 71.43\% & 48.25\% & \textbf{57.59\%} & 72.60\% & 46.49\% & 56.68\% \\
\hline

\end{tabular}
\end{center}
\end{table*}

\begin{figure}[!ht]
	\centering
	\includegraphics[width=0.4\textwidth]{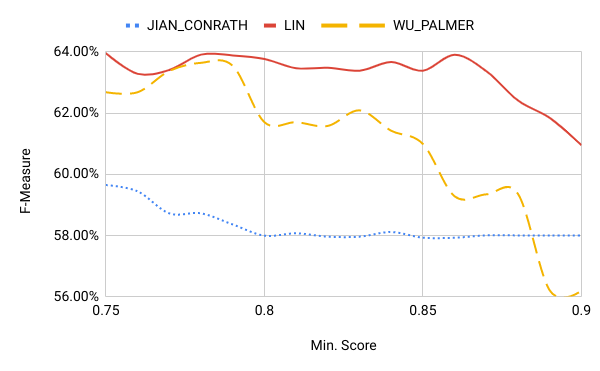}
    \caption{Variation of F-Measures with regard to Different Similarity Measures}
	\label{image:comparison_of_similarity_scores}
	\centering
\end{figure}

In addition to Wu-Palmer scores, we performed the same experiment on the verb pairs using all the eight semantic similarity measures available in Wordnet\cite{pedersen2004wordnet}. It was observed that Jiang-Conrath \cite{jiang1997semantic} and Lin \cite{lin1998information} are the two measures which provides reasonable accuracy in addition to Wu-Palmer semantic similarity\cite{wu1994verbs}. The results from these experiments are shown in Table \ref{table_different_scores} and in Fig.\ref{image:comparison_of_similarity_scores}. It could be observed that \textit{Lin}  outperforms other two measures when F-Measures are considered.
It can be seen that 0.75 is the \textit{Lin} score which has the highest F-Measure. But, it is due to considerably high recall and undesirably low precision values. As our intention is to maintain a proper balance between precision and recall, \textit{Lin} Score of 0.86 is selected as the threshold to detect verb pairs with similar meaning. 0.86 is the \textit{Lin} Score with the second highest F-Measure.





\subsection{Detecting Shift-in-View Relationships by Comparing Properties Related to Identified Verbs}

\subsubsection{Negation on Verbs}

Usage of negation relationship is a popular approach when it comes to detecting inconsistencies and contradictions in text \cite{marneffe2008finding,harris2009perspective,pubmed}. In this study, we checked for the negation relationship within verbs in verb pairs identified using the method proposed in the section 3.4. If one verb is detected as being negated while the other verb is not being negated, the sentence pair is considered as having \textit{Shift-in-View} relationship. Stanford CoreNLP dependency parser was used to detect the negation by identifying occurrences of the "neg" tag as described in "Stanford typed dependencies manual" \cite{de2008stanford}.   


\subsubsection{Using Adverbial Modifiers to Detect Shifts-In-View}

Another approach to detecting different viewpoints on the same subjects or entities can be formulated by considering adverbial modifiers. If the adverbial modifiers related to two verbs with similar meanings give opposite or contradictory meanings, that means the viewpoints on how that task was performed is different. Therefore, the adverbial modifiers related to the verbs in verb pairs identified using the methodology described in section 3.4 were considered. We classified adverbial modifiers in to three main classes shown in Table \ref{table:adverbial_modifier_list}. Within each class, there exists a positive subclass and a negative subclass. In the table, we have shown the positive sub classes with unshaded rows while the negative sub classes are shown with shaded rows.   
After defining major classes into which adverbial modifiers can be classified, lists containing adverbs related to each class were created. Table \ref{table:adverbial_modifier_list} further contains examples of adverbs related to each type. This table does not include all the adverbs we are maintaining in the lists.

\newcommand{\NegveS}{\cellcolor{gray!50!white}}

\begin{table}[h]
\caption{Adverbial Modifiers}
\label{table:adverbial_modifier_list}
\begin{center}
\begin{tabular}{|l|l|l|}
\hline
\textbf{Type Class} & \textbf{Type Name} & \textbf{Modifiers}\\
\hline
\multirow{2}{*}{Frequency} & more frequent & \begin{tabular}[c]{@{}l@{}}
 always, often, regularly
\end{tabular} 
\\
\hhline{~--}
& \NegveS less frequent & \NegveS 
\begin{tabular}[c]{@{}l@{}}
accidentally, never, not, less,
\\loosely, rarely, sometimes
\end{tabular}
\\
\hline
\multirow{2}{*}{Tone} & amplifiers &
\begin{tabular}[c]{@{}l@{}}
so, well, really, literally , simply,\\
for sure, completely, absolutely
\end{tabular}
\\
\hhline{~--}
& \NegveS down toners & \NegveS 
\begin{tabular}[c]{@{}l@{}}
kind of, sort of, mildly, \\ to some extent,
almost, all but
\end{tabular}
\\
\hline
\multirow{2}{*}{Manner} & positive manner & elegantly,beautifully,confidently\\
\hhline{~--}
& \NegveS negative manner & \NegveS lazily, ugly,faint heartedly\\
\hline

\end{tabular}
\end{center}
\end{table}

If adverbial modifiers connected to both verbs in a verb pair with similar meaning belong to same Adverbial modifier type, but with opposite polarities (one positive and one negative), it can be identified that the two sentences provide different views in relation to the entities that are connected by those verbs.  

\subsection{Discovering Inconsistencies between Triples}

Following the methodology presented in the PubMed study \cite{pubmed}, a legal term dictionary was constructed to be served as a \textit{Semantic Lexicon} for the system. 200+ legal opinion texts were used to extract words for the process. Then a word list consisting  17,000+ unique words were developed by removing stop words. A TF-IDF algorithm \cite{leskovec2014mining} based method is used to calculate a value for each term in the dictionary.

\begin{equation} 
\label{term_value}
Term Value = \frac{\sum_{i=1}^{case count} \frac{f_{t,d}}{term count}}{D.F}
\end{equation}

Raw count ($f_{t,d}$) for each term is taken, considering each legal opinion text as a seperate document. Term frequency value for a term is calculated by dividing the raw term count by the total number of terms in the case. Term frequency value for each case is added together and the result value is divided from the document frequency ($D.F$), to calculate the value for a term in the dictionary. Then all the term values are normalized according to the equation \ref{normalized_term_value}.

\begin{equation} 
\label{normalized_term_value}
NormalizedTV = \frac{(TV - TV_{min})*(1-TV_{min})}{TV_{max}-TV_{min}} + TV_{min}
\end{equation}

Here $TV_{min}$ and $TV_{max}$ represent the minimum and maximum values of the term values respectively. This normalized value is used to be served as the semantic weight for the system.

First, coreference resolving is done on the sentence pairs using the Stanford CoreNLP CorefAnnotator \cite{clark2015entity} and the pairs with \textit{Transition Words and Phrases} are filtered out.
Then OLLIE \cite{ollie-emnlp12}, open information extraction system, is used to extract triples, in (\textit{Subject; Relationship; Object}) format, from sentences. When comparing two sentences, for the \textit{Shift In View} relationship, only triple pairs with same subject or object are considered, as the \textit{Shift In View} relationship talks about different perspectives on the same topic or entity. The stop words removed relationship strings of a triple pair are then compared with each other word by word. The comparison is performed in three ways.

\begin{enumerate}
    \item Words which are exactly the same
    \item Exactly same words with one word negated with ``not"
    \item Different words
\end{enumerate}

In our study we consider the negation of words with similar meanings (\textbf{Lin} score above 0.86) instead of considering only the words which are exactly the same.
Then, an oppositeness value is obtained for each sentence pair by comparing the triples following the algorithmic approach proposed in the PubMed Study \cite{pubmed}. A threshold based on the oppositeness values is introduced empirically to select sentence pairs which have the \textit{Shift In View} relationship.


\subsection{Sentiment-based Approach}

Though valuable information can be obtained by analyzing the sentiment of a sentence, the sentiment of a sentence alone hardly gives any details on the topics which are being discussed within a sentence and on the viewpoint in which the sentence is describing the topic. It is known that the two sentences which are being compared to detect shifts in view discuss on the same topic as we consider only the sentence pairs with "Elaboration" relationship. But, when the sentences in legal opinion texts are considered, even if the sentiments of two sentences which elaborate on the same discussion topic is different, it can not be concluded that the two sentences are providing different opinions on the topic. 

The reason is that the person entities which are described in a sentence and connected with the sentiment of the sentence have a significant impact on the topic which is being discussed. For example, consider two sentences which elaborate on the same discussion topic and having opposite sentiments. If the sentiment of the sentence with negative sentiment is connected with the \textit{proposition} party while the sentiment of sentence with positive sentiment is connected with the \textit{opposition} party, it might be the case where both sentences are conveying opinions which are beneficial for the \textit{opposition} party in relation to the topic which is being discussed. 

The problem becomes even more complex when the sentence is made up of several sub-sentences because each sub-sentence may have a "Subject" of its own. Therefore, when using the sentiment based approach to detect "Shift-in-View" relationship, we consider only the sentence pairs in which each sentence has only one explicit subject. If the subjects in both sentences are the same in such a sentence pair, it can be concluded that two sentences are elaborating on the same topic in relation to the same subject. Then, it is checked whether the two sentences are providing sentiments with opposite polarities. If one sentence provides negative sentiment and other provides positive sentiment while discussing the same topic in relation to the same subject, it can be concluded that the probability of two sentences giving different perspectives on the same topic is very significant. 

In this approach, the sentences which are composed with subordinate clauses are first split using those clauses. When the sentence is split using a subordinating conjunction, that subordinate clause can be identified as another sentence entity. Throughout this section, we will refer the subordinate clause as \textit{inner sentence} and the main clause will be referred to as \textit{outer sentence}. After the sentence is annotated using Stanford CoreNLP Constituency Parser \cite{chen2014fast} , the splitting happens by identifying associated terms with \textit{SBAR} tag.  

The proposed approach is based on analyzing the sentiment of this inner sentence to identify if there is a \textit{shift in view }relation between a sentence pair. If we consider the Example 4 (which was taken from \textit{Lee v. United States} \cite{1977lee}), The phrases \textit{``Lee cannot convince the court that a decision to reject the plea bargain''}, and \textit{``he can establish prejudice under Hill''} are the inner sentences. The outer sentences are \textit{``The government argues''}, and \textit{``Lee, on the other hand, argues''}.
\begin{SentenceExample}
\item \begin{Sentence}The Government argues that Lee cannot "convince the court that a decision to reject the plea bargain.\end{Sentence}

\item \begin{Sentence}Lee, on the other hand, argues that he can establish prejudice under Hill.\end{Sentence}
\end{SentenceExample}

If we consider the sentence pair mentioned in Example 4, both the inner sentences' subject is Lee. The phrase \textit{``Lee cannot convince the court that a decision to reject the plea bargain''} is having a negative sentiment while the other inner sentence \textit{``Lee can establish prejudice under Hill''} denotes a positive sentiment. Both the outer sentences are having neutral sentiment. Therefore, it can be observed that there is a shift in view regarding the subject Lee.

%% file: body-sections/results.tex
\section{Experiments and Results}

 As the first step, the 3 major approaches used to detect \textit{Shift-in-View} relationship type were evaluated. In order to perform this evaluation, 2150 sentence pairs from legal opinion texts related to criminal court cases were extracted from Findlaw \cite{FindLaw}. Each of these sentence pairs contains two sentences which are consecutive to each other within a legal opinion text document. Next, the extracted sentence pairs were input into the Sentence Relationship Identifier (SRI). Input sentence pairs are first processed inside SRI. The sentence pairs which are identified as having \textit{Elaboration} by the SRI were further processed in order to detect whether there is \textit{Shift-in-View} relationship using the three \textit{Shift-in-View} detection approaches mentioned under Section III. 
 
 As the next step, the sentence pairs detected as having the \textit{Shift-in-View} relationship under each approach were taken into consideration. The number of detected sentence pairs from each approach is shown in Table \ref{table_results_comparison_both_agree}. Then, the precision of each approach was calculated. All 46 sentence pairs detected from Verb-Relationship approach were used when calculating the precision of that approach. 100 sentence pairs randomly selected from the detected 246 sentence pairs, which were identified using the Sentiment-Polarity approach was used to determine the precision of the approach. 95 sentence pairs were detected from the approach which uses inconsistencies between triples to determine \textit{Shift-in-View}. All of those 95 sentence pairs were used to calculate the precision of that approach. The precision values obtained for each of these approach are also shown in Table \ref{table_results_comparison_both_agree}. When performing this evaluation, each sentence pair was first annotated by two human judges. If the two judges did not agree on a relationship type for a particular sentence pair, that sentence pair was annotated by an additional human judge. When the results were calculated, the consideration was given only to the sentence pairs which were agreed by \textbf{at least two} human judges to have the same relationship type.
 
 Due to the scarcity of resources, it was not possible to annotate all 2150 sentence pairs based on the relationship type. As a result, calculating recall of each approach was not possible. If Table \ref{table_confusion_matrix} related to SRI study \cite{oblie2018identifying} is considered, it can be observed that only 3 out of 165 sentence pairs are determined as having the \textit{Shift-in-View} relationship type by the human judges. It suggests that the \textit{Shift-in-View} relationship type does not occur frequently when we consider sequential sentence pairs in a legal opinion text. Furthermore, Table \ref{table_confusion_matrix} suggests that the SRI tends to misattribute sentence pairs having \textit{Shift-in-View} as having the \textit{Elaboration} relationship type. That means, the SRI is successful in determining if the two sentences in the sentence pair is discussing the same topic or not. In such circumstances, it is important to be precise when determining a sentence pair as having the \textit{Shift-in-View} relationship type. Considering these facts, we can conclude that it is important to prioritize the \textit{Shift-in-View} detection approaches based on the precision.

\begin{table}[h]
\caption{Results Comparison of Approaches used to detect Shift-in-View}
\label{table_results_comparison_both_agree} 
\begin{center}
\begin{tabular}{|c|c|c|}
\hline
\thead{\bfseries Approach} & \thead{\bfseries No. of Sentence pairs} & \thead{\bfseries Precision}\\
\hline
Verb Relationships & 46 & 0.609 \\
\hline
Sentiment Polarity & 230 & 0.382 \\
\hline
Inconsistencies between triples & 95 & 0.273 \\
\hline

\end{tabular}
\end{center}
\end{table}

According to the Table \ref{table_results_comparison_both_agree}, it can be seen that the precision which could be obtained from analyzing relationships between verbs is around 0.6. As mentioned earlier we have selected the \textbf{Lin} semantic similarity score of 0.86 as the threshold to identify verbs with similar meaning after analyzing different semantic similarity measures. The precision of identifying verbs with 0.86 \textbf{Lin} score is 0.67. Thus, it can be seen that there is a potential to improve the  precision of detecting \textit{Shift-in-View} relationships using relationships between verbs by developing a semantic similarity measure which is more accurate in identifying verbs with similar meanings for the legal domain.

Using the sentiment based model, the achieved precision is 0.38. There are few possible reasons behind this observation. The study on the sentiment annotator model \cite{gamage2018Sentiment} used in this case, states that the accuracy of the model is 76\%. The study says that the errors present in its parent model \cite{socher2013recursive} can be propagated to the target model \cite{gamage2018Sentiment}. The paper on the source model\cite{socher2013recursive} which is based on recursive neural tensor network, shows that the accuracy is reduced down to 0.5 when the n-gram length of a phrase increases (n\textgreater10). As most of the sentences in court case transcripts are reasonably lengthier, there is a potential that the proposed sentiment based approach used for the identification of \textit{Shift-in-View} is affected by the above mentioned error. 

Only a precision of 0.27 could be observed in the approach which considers inconsistencies using triples as proposed in \textbf{PubMed Study}. The following reasons may have contributed to the poor performances of that approach. From the 2150 sentence pairs which were considered, oppositeness values were not calculated for 1570 pairs. Containing at least one sentence within a sentence pair in which the triples could not be extracted by OLLIE\cite{ollie-emnlp12} is a major reason for not having an oppositeness value. Even if the triples are extracted from both sentences, if there is no matching between either subjects or objects of the two sentences, an oppositeness value will not be calculated for a sentence pair. 

 Evaluation results demonstrates that analysis of relationships between verbs in two sentences as the only approach which performs the task of detecting \textit{Shift-in-View} relationships with a precision more than 0.5. Many studies convince the difficulty of detecting contradiction and change of perspective relationships  over other relationship types that can be observed between sentences\cite{marneffe2008finding,zahri2012exploiting, oblie2018identifying}. The study \cite{marneffe2008finding} also claims the difficulty of generalizing contradiction detection approaches. When considering these facts, it can be considered that the results obtained via analyzing verb relationships are satisfactory. Therefore, we combined only that approach with the Sentence Relationship Identifier (SRI) and evaluated the overall system made up by combining \textit{Shift-in-View} detection  with SRI as shown in Table \ref{table_results_5_classes}. 
 
 The results shown in Table \ref{table_results_5_classes} were obtained using 200 annotated sentence pairs. Each of the considered sentence pair was agreed by at least two human judges to have the same relationship type. Furthermore, 21 randomly selected sentence pairs which were agreed by at least two human judges as having \textit{Shift-in-View} are contained within the 200 sentence pairs which were used in this evaluation.
 
 \begin{table}[h]
\caption{Results Obtained from Sentence Pairs in which At least Two Judges Agree}
\label{table_results_5_classes} 
\begin{center}
\begin{tabular}{|c|c|c|c|}
\hline
\thead{\bfseries Discourse Class} & \thead{\bfseries Precision} & \thead{\bfseries Recall} & \thead{\bfseries F-Measure} \\
\hline
Elaboration & 0.938 & 0.930 & 0.933\\
\hline
No Relation & 0.843 & 0.895 & 0.868\\
\hline
Citation & 1.000 & 0.971 & 0.985\\
\hline
Shift-in-View & 0.688 & 0.423 & 0.524\\
\hline

\end{tabular}
\end{center}
\end{table}

According to the Table \ref{table_results_5_classes}, it can be seen that there is a significance improvement, especially in relation to the \textit{Shift-in-View} relationship type when compared with the results in the study\cite{oblie2018identifying} as given in Table \ref{table_results_comparison_both_agree_original}.

%% file: body-sections/conclusions.tex
\section{Conclusion}

Developing a methodology to detect situations where multiple viewpoints are provided in regard to the same discussion topic within a legal opinion text is the major research contribution of this study. This study has introduced novel approaches to detect deviations in the opinions provided by two sentences regarding the same topic. At the same time, existing methodologies to detect contradiction and change of perspectives have been evaluated within the study. Additionally, it has been empirically demonstrated the way in which the outcomes of the study can be used to facilitate the process of identifying relationships between sentences in documents containing legal opinions on court cases. Evaluation of the performance of existing semantic similarity measures in relation to identifying verbs with similar meaning can be considered as another key research contribution of the study.    

The proposed approach can also be used to facilitate several other information extraction tasks related to the legal domain such as identifying counter arguments to a particular argument, determining representatives related to the proposition party and the opposition party in a court case.

The accuracy of the approaches proposed in this study can be further improved by developing semantic similarity measures and sentiment annotators which can perform in the legal domain with an improved accuracy. Coming up with such mechanisms can be considered as the major future work.